\documentclass[conference, 9pt]{IEEEtran}
\IEEEoverridecommandlockouts

\usepackage{cite}
\usepackage{amsmath,amssymb,amsfonts}
\usepackage{algorithmic, algorithm}
\usepackage{graphicx}
\usepackage{textcomp}
\usepackage{xcolor}
\usepackage{amsthm}
\usepackage{multirow}
\usepackage{enumitem}
\usepackage{subcaption}
\usepackage{color}
\usepackage{caption}
\usepackage{needspace}

\usepackage{array}

\usepackage[caption=false,font=normalsize,labelfont=sf,textfont=sf]{subfig}
\usepackage[margin=0.75in]{geometry}
\usepackage{stfloats}
\usepackage{url}
\usepackage{lineno}
\usepackage{verbatim}
\usepackage{balance}
\usepackage{subcaption}

\def\BibTeX{{\rm B\kern-.05em{\sc i\kern-.025em b}\kern-.08em
    T\kern-.1667em\lower.7ex\hbox{E}\kern-.125emX}}

\begin{document}
\IEEEpubid{\makebox[\columnwidth]{979-8-3315-3762-3/25/\$31.00~\copyright~2025 IEEE\hfill}\hspace{\columnsep}\makebox[\columnwidth]{}} 
\IEEEpubidadjcol

\title{Recursive Learning-Based Virtual Buffering for Analytical Global Placement}
\author{
Andrew B. Kahng,
Yiting Liu,
and Zhiang Wang \\
\{abk,yil375,zhw033\}@ucsd.edu
\medskip
\\
\small
UC San Diego, La Jolla, CA, USA \\
\thanks{Revised version, November 2026. This revision corrects a unit-conversion issue in the released
{\em MLBuf-RePlAce} integration~\cite{Issue6}, replaces the buffering model by the retrained {\em MLBuf v2}
(Section~\ref{sec:MLBuf}), and re-runs all experiments 
(Section~\ref{sec:experiments}). The {\em MLBuf-RePlAce} and Ad-Hoc results of the MLCAD 2025 version~\cite{KahngLW25} are
superseded by the results in this revision.}
}

\maketitle

\begin{abstract}

With scaling of interconnect versus gate 
delays in advanced technology nodes,
placement with {\em buffer porosity} 
awareness is essential for timing closure
in physical synthesis flows. 
However, existing approaches face two key challenges: 
(i) traditional van Ginneken-Lillis-style buffering 
approaches \cite{LillisCL96, VanGinneken90} 
are computationally expensive during global placement;
and (ii) machine learning-based approaches, such as {\em BufFormer}~\cite{LiangNRHR23}, 
omit important Electrical Rule 
Check (ERC) considerations and typically fail to ``close the loop'' 
back into the physical design flow.
In this work, we propose {\em MLBuf-RePlAce}, 
an open-source learning-driven virtual 
buffering-aware analytical global placement framework, 
built on top of the OpenROAD infrastructure \cite{OpenROAD}.
{\em MLBuf-RePlAce} adopts an efficient recursive 
learning-based generative buffering approach 
to predict buffer types and locations, addressing 
ERC violations during global placement.
We compare {\em MLBuf-RePlAce} against the default 
virtual buffering-based timing-driven global placer 
in OpenROAD, using open-source testcases from the 
TILOS MacroPlacement \cite{MacroPlacement} and 
OpenROAD-flow-scripts \cite{ORFS} repositories.
Without degradation of post-route power, 
{\em MLBuf-RePlAce} achieves (maximum, average) 
improvements of (49\%, 24\%) in total negative slack 
(TNS) and (38\%, 19\%) in worst negative slack (WNS)
within the open-source OpenROAD flow.
When evaluated by completion in a commercial flow
under relaxed timing constraints, our approach achieves
(maximum, average) improvements of (83\%, 69\%) in TNS.
\end{abstract}


\section{Introduction}
\label{sec:introduction}

Global placement is a critical step in VLSI physical design.
The quality and efficiency of global placement significantly 
impacts the timing closure of the place-and-route (P\&R) flow.
State-of-the-art analytical global placers 
\cite{ChengKKW19}\cite{DuGLWH24}\cite{GuthLNFGS15}\cite{LinWCZC24} 
typically adopt the electrostatics-based placement approach~\cite{LuCCLH15}, 
formulating global placement
as nonlinear programming under density constraints.
GPU-accelerated placers~\cite{LiaoLCLLY22}
\cite{KahngW25}\cite{GuoL22} parallelize the computation 
of wirelength and density functions, achieving significant 
speedups. 
Additionally, for design implementation in advanced technology nodes, 
timing closure requires 
extensive buffer insertion \cite{LiangNRHR23} and brings a complex
interplay with global placement.
Placement with {\em buffer porosity} 
awareness~\cite{ChenCP08} is needed to allocate enough 
space for buffer insertion at later optimization stages; 
typically, this requires invocation of a buffering engine multiple 
times during global placement.

Traditional van Ginneken-Lillis-style buffering 
approaches \cite{LillisCL96}\cite{VanGinneken90} 
typically begin with constructing a Steiner 
minimum tree \cite{ChuW08} or a timing-driven 
tree \cite{AlpertHHKL01}\cite{AlpertGHHQS04}, 
followed by a bottom-up dynamic programming 
process to determine buffer types and locations.
However, the joint space of tree generation 
and buffer insertion can be huge \cite{CongY00}, 
especially for nets with many sinks,
making such approaches computationally 
expensive during global placement.
Moreover, in a GPU-accelerated placer, actual 
buffer insertion changes the 
netlist topology, which requires updates 
to the netlist stored in GPU memory. Frequent memory 
accesses and updates significantly degrade the speed 
advantages provided by GPU acceleration. 
Therefore, it is essential to develop a \textbf{lightweight
virtual buffering strategy} to guide the placer
toward a buffer-porosity-aware solution
without compromising computational efficiency.

Machine learning (ML)-based buffering approaches 
have recently attracted considerable attention in the research literature.
Wu et al.\cite{WuHLZ24} propose a 
reinforcement-learning (RL)-based approach 
for simultaneous gate sizing 
and buffer insertion.
Liang et al.\cite{LiangNRHR23} introduce 
{\em BufFormer}, a generative machine learning framework 
targeting buffering at the Engineering 
Change Order (ECO) stage.
While both approaches show promise in delay optimization, 
they share two limitations:
(i) they lack a thorough consideration of 
Electrical Rule Check (ERC) 
(maximum slew, maximum capacitance and 
maximum fanout) constraints, and 
(ii) they do not ``close the loop'' 
back to evaluate in the full physical design flow.

In this work, we present a recursive learning-based 
virtual buffering framework for addressing 
ERC violations in VLSI placement.
We further integrate the proposed generative ML framework
into a high-quality open-source electrostatic-based 
global placer, enabling placement with buffer porosity awareness~\cite{AlpertGHHQS04}\cite{ChenCP08}.
Our main contributions are as follows.
\begin{itemize}[noitemsep, topsep=0pt, leftmargin=*]

\item We propose {\em MLBuf}, a recursive 
learning-based generative buffering model
that efficiently predicts buffer types and locations 
to address ERC violations.
{\em MLBuf} employs a recursive strategy 
inspired by the bottom-up dynamic programming paradigm 
used in the classical van Ginneken-Lillis-style algorithms~\cite{LillisCL96}\cite{VanGinneken90}. 
Additionally, a differentiable clustering approach
is proposed to avoid Steiner tree construction,
enabling the exploration of more types of buffer-embedded trees.

\item We develop {\em MLBuf-RePlAce}, 
a learning-driven virtual buffering-aware 
analytical global placement framework. 
To the best of our knowledge, we are the first 
to explore and assess ML-guided buffering
for analytical placement in the context of 
the ``full placement and optimization flow''.
{\em MLBuf-RePlAce} is built on the OpenROAD 
infrastructure with a permissive open-source license, 
enabling others to 
adapt it for future enhancements.

\item 
We evaluate {\em MLBuf-RePlAce} framework using 
both OpenROAD and commercial flows, 
along with open testcases from the TILOS MacroPlacement~\cite{MacroPlacement} and OpenROAD-flow-scripts \cite{ORFS} GitHub repositories.
We compare our approach against the default 
virtual buffering-based timing-driven global placer 
in the OpenROAD flow.
Within the open-source OpenROAD flow, our approach
achieves (maximum, average) improvements of (49\%, 24\%)
in TNS. When evaluated by completion in a commercial flow
under relaxed timing constraints, our approach achieves
(maximum, average) improvements of (83\%, 69\%) in TNS.

\end{itemize}

The remaining sections are organized as follows. 
Section \ref{sec:preliminary} introduces the terminology 
and background. Section \ref{sec:approach} and 
Section \ref{sec:MLBuf} discuss our approach. 
Section \ref{sec:experiments} shows experimental results,
and Section \ref{sec:conclusion} concludes the paper.

\section{Preliminaries}
\label{sec:preliminary}

This section reviews the nonlinear analytical 
global placement method and presents the problem 
formulation for ML-based buffer insertion. 
All terms used in this paper are summarized in
Table~\ref{tab:notations}.

\begin{table}[!htb]
\caption{Terminology and Notation.}
\label{tab:notations}
\centering
\scriptsize
\begin{tabular}{|p{2cm}|p{5.5cm}|}
\hline
\multicolumn{2}{|c|}{\textbf{Basic Entities}} \\ \hline
$v$       & Cell (driver, sink or buffer) \\ \hline
$p$       & Cell pin or input-output pin \\ \hline
$e$        & Net $e = \{p\}$\\ \hline
$V$       & Set of all cells $\{v\}$ \\ \hline
$\mathcal{V}^s$       & Set of drivers and sinks \\ \hline
$\mathcal{V}^b$       & Set of buffers \\ \hline
$E$        & Set of all nets $\{e\}$\\ \hline
$x_v$     & Cell location, $\forall v \in V$ \\ \hline
$N$       & Number of cells (driver and sinks) in one net \\ \hline

\multicolumn{2}{|c|}{\textbf{Electrical Properties}} \\ \hline
$C_{in}(p)$   & Input capacitance of pin $p$ \\ \hline
$C_{out}(p)$  & Output capacitance of pin $p$ \\ \hline
$C_{max}(p)$  & Max (load) capacitance of driver pin $p$ \\ \hline
$S_{in}(p)$   & Input slew of pin $p$ \\ \hline
$S_{out}(p)$  & Output slew of pin $p$ \\ \hline
$O(p)$        & Fanout of driver pin $p$ \\ \hline
$O_{max}(p)$  & Max fanout of driver pin $p$ \\ \hline
$R$           & Resistance \\ \hline

\multicolumn{2}{|c|}{\textbf{Placement and Geometry}} \\ \hline
$\mathcal{W}$       & Half-perimeter wirelength \\ \hline
$\mathcal{W}_{max}$ & Max wirelength constraint \\ \hline
$b$                 & Rectangular bin in placement region \\ \hline
$\mathcal{B}$       & Set of all bins $\{b\}$ \\ \hline
$h$                 & Bounding box of the buffer \\ \hline
$\mathcal{H}$       & Set of buffer bounding boxes $\{h\}$ \\ \hline
$A$                 & Area of cells or bins \\ \hline
$A_{\text{grid}}(b_j)$ & Total area allowed in bin $b_j$\\ \hline
$A_{\text{cell}}$   & Cell area function in bins \\ \hline
$A_{\text{cell}}^{\prime}$   & Total area of movable cells placed in a bin \\ \hline
$A_{\text{overlap}}(h_i, b_j)$   & Overlapping area of the buffer $h_i$ and the bin $b_j$\\ \hline
$D_j$               & Target density of bin $b_j$ \\ \hline
$\tilde{H}$         & Level of the buffer-embedded tree \\ \hline

\multicolumn{2}{|c|}{\textbf{Model Features and Embeddings}} \\ \hline
$F \in \mathbb{R}^{n \times m}$ & Feature matrix \\ \hline
$\tilde{x}$ & Feature embedding \\ \hline
$d$ & Feature dimension \\ \hline
$K$ & Cluster number \\ \hline
$M_k$ & Assignment matrix with $k$ clusters \\ \hline
$\tilde{C}$ & Cosine distance between embeddings \\ \hline
$g$ & Gumbel noise \\ \hline
$Z^{type}$ & Predicted buffer types \\ \hline
$Z^{loc}$  & Predicted buffer locations \\ \hline
$X$ & Number of buffer types \\ \hline
$\oplus$ & Embedding concatenation \\ \hline
$c$ & Net-context feature vector \\ \hline
$\hat{g}$ & Predicted probability that a net needs buffering (gate) \\ \hline
$m$ & Soft member count of each cluster \\ \hline
$\Phi_c$ & Cluster-physics features (member count, capacitance, bounding box, HPWL, load) \\ \hline
$\pi_c$ & Load-table prior over buffer types \\ \hline

\multicolumn{2}{|c|}{\textbf{Learning and Training Terms}} \\ \hline
$L$ & Loss functions \\ \hline
$P$ & Penalty terms \\ \hline
$W$ & Weight of loss and penalty \\ \hline
$f_\theta$ & Fully connected networks with parameter $\theta$ \\ \hline
$t_\zeta$ & Self-attention layer with parameter $\zeta$, \\ \hline
$\mathcal{F}_\delta$ & MLBuf model with parameter $\delta$,
$\mathcal{F}_\delta = \text{Stack}(t_{\zeta}, f_{\theta})$\\ \hline
$\alpha$ & Scaling factor for wirelength estimation \\ \hline
$\beta$  & Fitting factor for output slew \\ \hline
 \{$\mathbf{X}_i, \mathcal{Y}_i$\} & Input-label pairs in ground truth buffer-embedded tree \\\hline
 $T$ & Ground truth buffer-embedded tree in training dataset \\\hline
 $\hat{T}$ & Predicted buffer-embedded tree during model inference\\\hline
 $\mathcal{D}_{train}$ & Training dataset \\\hline
\end{tabular}
\vspace{-1.5em}
\end{table}

\subsection{Nonlinear Analytical Global Placement}
\label{subsec:global_placement}

State-of-the-art global placers, such as 
{\em RePlAce}~\cite{ChengKKW19}  
and {\em DREAMPlace} \cite{LiaoLCLLY22}, 
usually adopt the 
electrostatics-based placement approach~\textcolor{black}
{\cite{LuCCLH15}}. 
Let $V$ = \{$v_1$, $v_2$, ..., $v_n$\} 
represent cells and $E$ = \{$e_1$, $e_2$, ..., $e_m$\}
represent nets. Let $x_v$ denote the coordinates 
of the center of cell $v$.
Electrostatics-based global placers dissect the placement region 
into a grid of non-overlapping bins and optimize wirelength 
under the density constraint:
\begin{equation}
\begin{aligned}
  & \text{min} \quad \mathcal{W}(V, E; x_v) \\
   & \text{s.t.} \quad A_{\text{cell}}(V; x_v) \leq A_{\text{grid}}, \quad \text{for each bin in the bin } \text{grid}
\end{aligned}
\end{equation}
where $\mathcal{W}(V, E ; x_v)$ is the wirelength function,
{$A_{\text{cell}}(V; x_v)$} is the cell area function in bins,
and $A_{\text{grid}}$ is the total area allowed in a bin.
In {\em MLBuf-RePlAce} framework, 
we dynamically adjust $A_{\text{grid}}$ based on buffering results
during the global placement process.

\subsection{Problem Formulation}
\label{subsec:problem_formulation}

The net-level buffering problem during global placement 
is defined as follows.
We are given: 
\begin{itemize}[noitemsep, topsep=0pt, leftmargin=*] 

\item a driver pin, its location and 
input slew,\footnote{The input slew is provided by 
OpenSTA~\cite{OpenSTA}.} 

\item a set of sink pins and their locations, 

\item a buffer library with associated cell information, 
including area, input capacitance, output resistance 
and maximum output capacitance, 

\item the input capacitance of each sink, 

\item the output resistance of the driver, 

\item electrical properties of interconnects for 
estimating resistance and capacitance,

\item electrical rule check (ERC) constraints for 
each cell (including buffers), 
i.e., maximum slew (max\_slew), maximum 
capacitance (max\_cap) 
and maximum fanout (max\_fanout) constraints, and

\item a maximum wirelength constraint (max\_wirelength).

\end{itemize}

The objective is to construct a \emph{buffer-embedded 
tree}, i.e., a tree with inserted buffer 
types and locations, that minimizes total buffer area
while satisfying all ERC and maximum wirelength constraints.


\section{Our approach}
\label{sec:approach}

\begin{figure}[htp]
\includegraphics[scale=0.32]{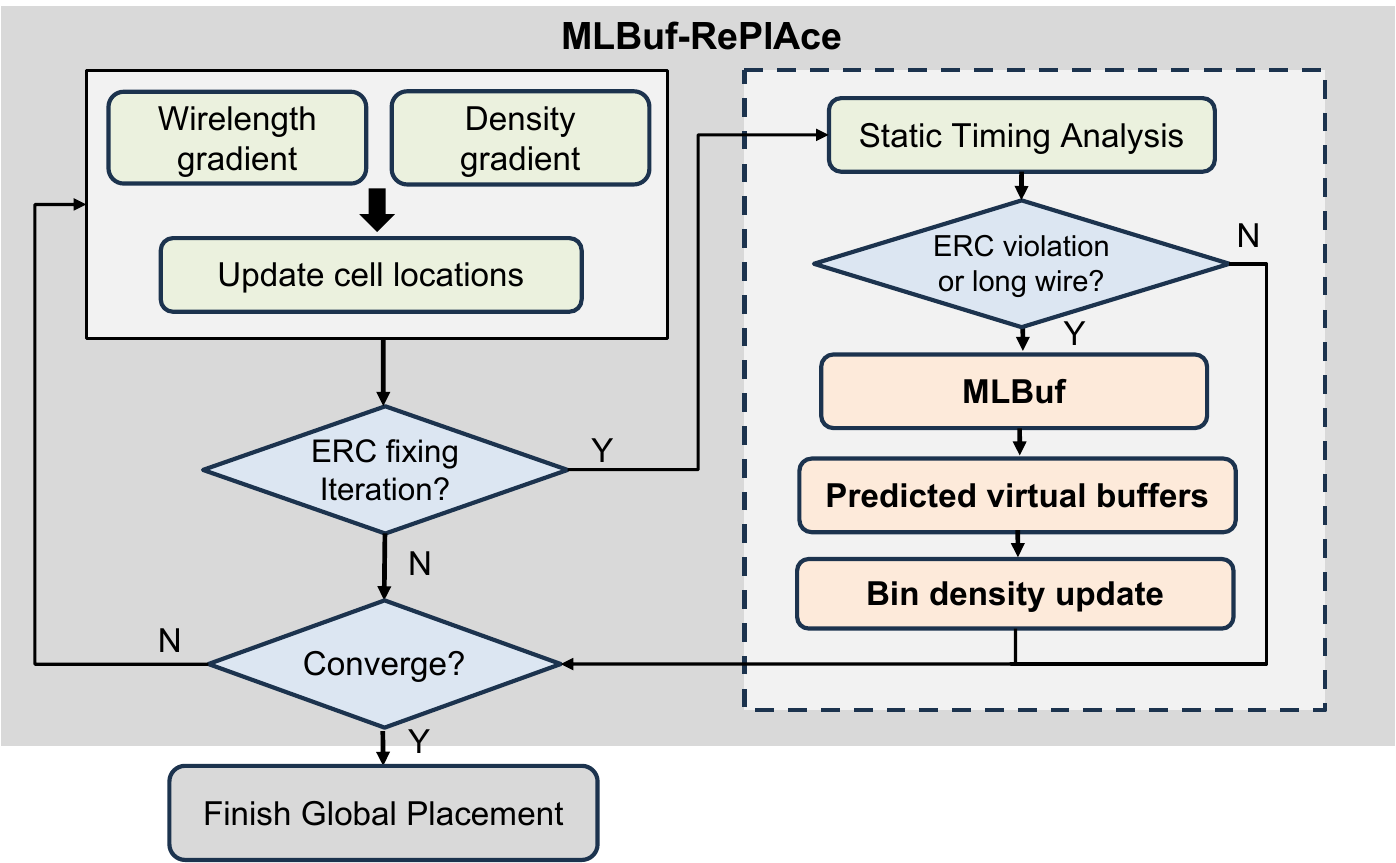}
\centering  
\caption{The flow of {\em MLBuf-RePlAce}.}\label{fig:overall_flow}
\vspace{-1em}
\end{figure}

Figure \ref{fig:overall_flow} shows our approach.
During timing-driven global placement, 
when the placement overflow reaches 
a threshold specified in a predefined 
overflow\_list,
a static timing engine (OpenSTA \cite{OpenSTA} in this work) 
is invoked to identify 
problematic nets with ERC violations \cite{gpl}.
Problematic nets, which also include long nets, are 
fed to the pre-trained {\em MLBuf} model for 
repair of ERC violations. (Details of {\em MLBuf} are 
given in Section \ref{sec:MLBuf}.)
{\em MLBuf} generates the buffer-embedded tree 
(with buffer types and locations) for each problematic net. 
Then {\em virtual buffering} is performed by 
pre-allocating {\em virtual occupancy} within placement 
bins according to the generated buffer-embedded trees. 
This pre-allocation allows the global placer 
to reserve adequate space for buffer insertion during 
later place-and-route stages, thus mitigating 
routing congestion while improving post-route power and timing.

More specifically, the virtual occupancy is calculated as follows.
Let $\mathcal{H} = \{h_1, \dots, h_i, \dots,  h_n\}$ denote 
predicted bounding boxes of virtual buffers, and
$\mathcal{B} = \{b_1, \dots, b_j, \dots, b_m\}$ denote 
bins in the placement region.
The overlapping area of $(h_i, b_j)$ is calculated as
\begin{equation}
\begin{aligned}
A_{\text{overlap}}(h_i, b_j) =& \max(0, \min(h_i^{rx}, b_j^{rx}) - \max(h_i^{lx}, b_j^{lx})) \\
&\times \max(0, \min(h_i^{ry}, b_j^{ry}) - \max(h_i^{ly}, b_j^{ly})),
\end{aligned}
\end{equation}

\noindent where $(h_i^{lx}, h_i^{ly})$ and $ (h_i^{rx}, h_i^{ry})$ 
denote the lower-left and upper-right coordinates 
of the bounding box for the predicted buffer $h_i$,
and similarly for bin $b_j$.
The total buffer-induced area consumption in bin $b_j$ is then scaled by its target density $D_j$, and subtracted from the available bin area:
\begin{equation}
A_{\text{grid}}(b_j)  \mathrel{-}=A_{\text{overlap}}(h_i, b_j) \cdot D_j.
\end{equation}
\noindent
The resulting overflow for bin $b_j$ is calculated as:
\begin{equation}
\text{overflow}(b_j) = \max\left(0, A_{\text{cell}}^{\prime}(b_j) - A_{grid}(b_j)\right),
\end{equation}

\noindent where $A_{\text{cell}}^{\prime}(b_j)$ is the total area 
of movable cells placed in bin $b_j$. 
This formulation of virtual occupancy allows the global placer to account for
the required buffer porosity,
guiding cell distribution accordingly during global placement.



\section{MLBuf: Recursive Learning-based Buffering}
\label{sec:MLBuf}

This section introduces the recursive 
learning-based generative buffering model {\em MLBuf}.
{\em MLBuf} mimics bottom-up dynamic programming  
to hierarchically construct the buffer-embedded 
tree for fixing ERC violations.
We formulate the buffer insertion process as 
a sequence transformation problem.
Each net is considered independently. Cells within
the same net are regarded as a sequence with associated
spatial and electrical information.
{\em MLBuf} iteratively captures key features from input
sequences to predict transformed sequences that include not
only the original drivers and sinks, but also the inserted
buffers. The output of each iteration is fed into 
the next iteration, enabling a hierarchical construction 
of buffer-embedded trees.
Observe that no Steiner tree 
construction is needed when applying {\em MLBuf}.

\begin{figure}[htbp]
\includegraphics[scale=0.28]{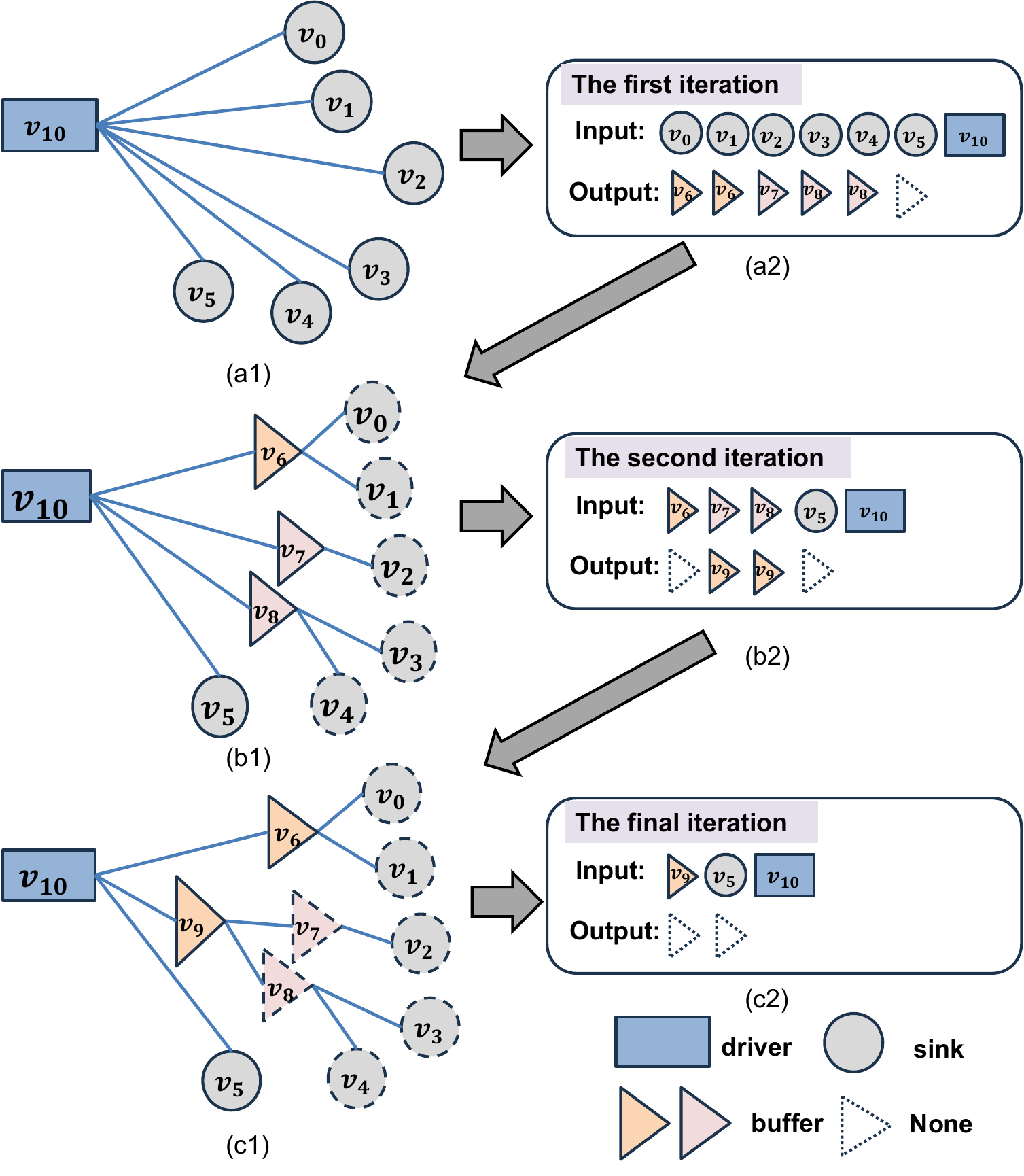}
\centering  
\caption{{\em MLBuf}'s bottom-up 
learning process.}\label{fig:learning_process}
\vspace{-0.5em}
\end{figure}

Figure~\ref{fig:learning_process} illustrates {\em MLBuf}'s 
bottom-up learning process.
In the first iteration (Figure~\ref{fig:learning_process}(a1) 
and Figure~\ref{fig:learning_process}(a2)), the input sequence 
is the original net with 
the driver $v_{10}$ and all sinks $v_0, \dots, v_5$.
{\em MLBuf} determines which sinks need to be merged 
and driven by the same buffer, and output their
corresponding ``parent'' buffers
with buffer types and locations.
In this example, sinks $v_0$ and $v_1$ are merged
and driven by the buffer $v_6$; sink $v_2$ is driven
by the buffer $v_7$; sinks $v_3$ and $v_4$ are 
merged and driven by the buffer $v_8$; and sink $v_5$
is not buffered, which is indicated by the 
corresponding buffer type None.
As buffers shield the effects of downstream
sinks, the input for the subsequent iteration 
(see the input in Figure~\ref{fig:learning_process}(b2)) 
contains the driver ($v_{10}$), 
buffers predicted in the last iteration ($v_6, v_7$ and $v_8$), 
and any remaining sinks that were not buffered in the last 
iteration ($v_5$).
The iterations end when all predicted buffer
types are None, indicating no further buffering is needed
(see the output in Figure~\ref{fig:learning_process}(c2)).
Then, the entire buffer-embedded tree is constructed 
(see Figure~\ref{fig:learning_process}(c1)).

\noindent\emph{MLBuf introduces three key innovations, as follows.}

First, constructing hierarchical buffer-embedded 
trees requires clustering of sinks to guide tree topology. 
Traditional clustering methods involve discrete operations
that disrupt gradient propagation and prevent end-to-end 
optimization. To address this issue, we introduce a 
\textbf{differentiable clustering module} that enables 
end-to-end training, effectively bridging the gap between 
discrete clustering operations and continuous neural network 
optimization (see Section~\ref{subsec:model_structure}).

Second, the recursive structure of {\em MLBuf} can
lead to error accumulation across iterations, 
potentially resulting in a buffer-embedded
tree with low accuracy.
To mitigate this issue, 
we adopt the \textbf{teacher forcing strategy}~\cite{WilliamsZ89} 
during training, which helps stabilize the 
learning process and reduce error propagation 
across recursive iterations
(see Section~\ref{subsec:training_strategy}).

Third, in addition to predicting buffer types and locations,
{\em MLBuf} also needs to determine whether a net needs buffering.
We introduce a \textbf{net-level gate} trained with unbuffered
nets as negative samples. We further propose \textbf{set-based losses} to supervise buffer
types, locations, counts and spatial coverage using one-to-one
Hungarian matching~\cite{Kuhn55}~\cite{CarionMSUKZ20}, count loss and Chamfer loss~\cite{FanSG17} 
(Section~\ref{subsec:training_loss}).

\begin{figure*}[htbp]
\includegraphics[scale=0.45]{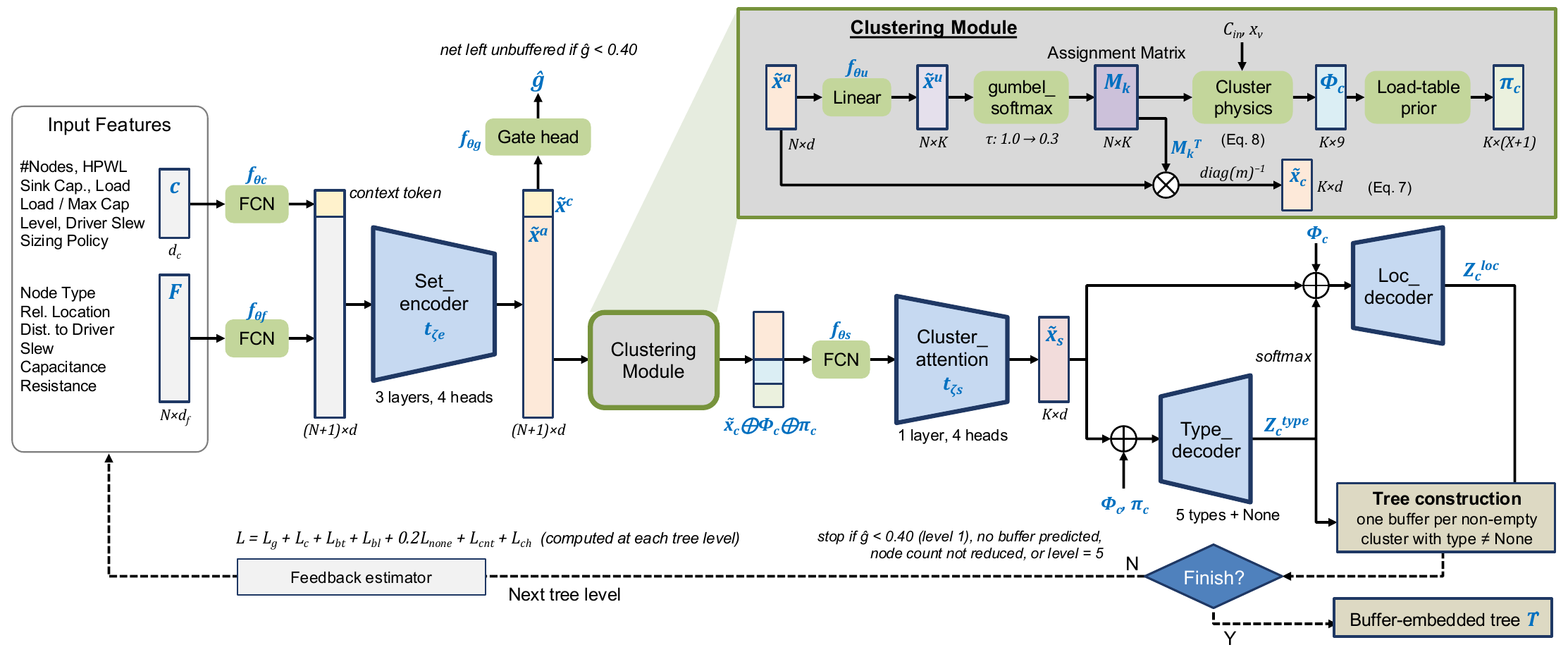}
\centering  
\caption{Model structure of {\em MLBuf}.}
\label{fig:model}
\vspace{-1em}
\end{figure*}

\subsection{Model Structure}
\label{subsec:model_structure}

Figure~\ref{fig:model} illustrates the structure of {\em MLBuf}.
The model consists of three components: (i) a set encoder with
a net-context token; (ii) a differentiable clustering module
with cluster-level electrical and spatial information; and
(iii) two task-specific decoders.
The set encoder extracts spatial and electrical features of these nodes,
together with net-level information. 
A net-level \emph{gate}
predicts whether buffering is needed. 
The clustering module
groups nodes that may be driven by the same buffer.
Finally, \emph{Type\_decoder} predicts the buffer type of each
cluster, and \emph{Loc\_decoder} predicts its location.

\noindent\textbf{Set Encoder: }
The input of the set encoder consists of a node feature
matrix $F \in \mathbb{R}^{N \times d_f}$ ($d_f = 15$), with
one row per node, and a net-context vector
$c \in \mathbb{R}^{d_c}$.
Detailed features are listed in
Appendix~\ref{subsec_app:model_details}.
Two-layer fully-connected networks (FCNs) project the node
and net-context features into $d$-dimensional embeddings.
The net-context embedding is then concatenated to the node
embeddings as a context token. A three-layer, four-head
transformer encoder~\cite{VaswaniSPUJGKP}
$t_{\zeta_e}$ processes the resulting sequence:
\begin{equation}\label{eq:encoder}
[\tilde{x}^c;\, \tilde{x}^a] = t_{\zeta_e}\!\left([f_{\theta_c}(c);\, f_{\theta_f}(F)]\right),
\quad \tilde{x}^a \in \mathbb{R}^{N \times d},
\end{equation}
where $\tilde{x}^c \in \mathbb{R}^{d}$ and
$\tilde{x}^a \in \mathbb{R}^{N \times d}$ denote the context
and node embeddings, respectively. The self-attention layers
capture dependencies among nodes of the same net, while the
context token provides net-level information.
A gate head $f_{\theta_g}$ predicts the probability
$\hat{g} = \sigma(f_{\theta_g}(\tilde{x}^c))$ that the
net requires buffering, where $\sigma$ is the logistic
function. Nets with $\hat{g}$ below the validation-selected
threshold of 0.40 are left unbuffered.

\noindent\textbf{Clustering Module: }
To avoid expensive Steiner tree construction,
{\em MLBuf} employs a differentiable clustering module
that partitions the input nodes into $K$ candidate clusters.
Nodes assigned to the same cluster are jointly considered
for buffer insertion. The driver is not clustered and remains
the root of the buffer-embedded tree.
A linear layer maps each node embedding to $K$ assignment
scores, $\tilde{x}^u = f_{\theta_u}(\tilde{x}^a)
\in \mathbb{R}^{N \times K}$.
To enable end-to-end training,
a soft cluster assignment matrix $M_k \in \mathbb{R}^{N \times K}$
is generated using Gumbel-Softmax~\cite{JangGP17},
\begin{equation}\label{eq:gumbel_softmax}
[M_k]_{ij} = \frac{\exp\left((\tilde{x}^u_{ij} + g_{ij}) / \tau\right)}{\sum_{q=1}^{K} \exp\left((\tilde{x}^u_{iq} + g_{iq}) / \tau\right)},
\end{equation}
where $g_{ij} \sim \text{Gumbel}(0, 1)$ is
Gumbel noise, and $\tau$ is a temperature
parameter that controls the ``sharpness''
of the distribution. As $\tau \rightarrow 0$,
the distribution becomes closer to a one-hot vector.
This relaxation allows the model to learn
cluster assignments in a fully differentiable
manner.
The temperature is annealed from 1.0 to 0.3
during training. During inference, Gumbel noise is removed
and each node is assigned to the cluster with the maximum
score.
The resulting matrix $M_k$ serves as a soft
indicator of cluster membership: each row
$i$ corresponds to a cell, and each column $j$
represents the probabilities of cells being assigned
to cluster $j \in K$.
The soft assignment matrix $M_k$ is used to
(i) aggregate node embeddings into cluster-level
representations through mass-normalized pooling
(Eq.~\ref{eq:cmodule_clevel_rep}), and
(ii) calculate differentiable cluster-level electrical
and spatial features (Eq.~\ref{eq:cluster_physics}).

The cluster-level representations are calculated as
\begin{equation}\label{eq:cmodule_clevel_rep}
\tilde{x}_c = \mathrm{diag}(m)^{-1} M_k^T \tilde{x}^a  \in \mathbb{R}^{K \times d},
\qquad m = M_k^T \mathbf{1}_N \in \mathbb{R}^{K},
\end{equation}
where $m$ denotes the soft number of nodes assigned
to each cluster. The cluster-physics features $\Phi_c \in \mathbb{R}^{K \times 9}$
collects nine values per cluster: the member count
$m$; total input capacitance $C_c = M_k^T C_{in}$; the center
coordinates $(c_x, c_y)$, width and height of a soft bounding box
of the members, where $(c_x, c_y)$ is the $M_k$-weighted mean of
the member locations; the half-perimeter $\mathcal{W}_c$ of that
box; the estimated load $C_{load}(c)$ of Eq.~(\ref{eq:cluster_physics});
and a non-empty indicator that marks the
clusters that have at least one member.
The estimated load is
\begin{equation}\label{eq:cluster_physics}
C_{load}(c) = C_c + \alpha\, \mathcal{W}_c \times C_{wire\_unit}.
\end{equation}
Based on $C_{load}(c)$, we construct a \emph{load-table prior}
$\pi_c \in \{0,1\}^{K \times (X+1)}$, which is a one-hot encoding
of the buffer type selected by a load-based sizing rule.
The rule selects BUF\_X2 for loads up to 11.5\,fF,
BUF\_X4 up to 30.1\,fF, BUF\_X8 up to 49.2\,fF,
BUF\_X16 up to 133\,fF, and BUF\_X32 for larger loads.
It reproduces 86\% of OR rsz's buffer-type selections in
the training data. We provide $\Phi_c$ and $\pi_c$ 
to the decoders, allowing them to use
ERC-related information rather than inferring it solely
from node features.

\noindent\textbf{Decoders: }
The cluster embeddings, cluster-physics features
and load-table prior are concatenated and projected to $d$
dimensions. A self-attention layer $t_{\zeta_s}$ then models
the dependencies among the $K$ clusters
$\tilde{x}_s = t_{\zeta_s}(f_{\theta_s}[\tilde{x}_c \oplus \Phi_c \oplus \pi_c])
\in \mathbb{R}^{K \times d}$.
The resulting cluster representations $\tilde{x}_s$ are the input of
the type and location decoders below.
The buffer type decoder, \emph{Type\_decoder}, predicts
the buffer type distribution
$Z_{c}^{type} \in \mathbb{R}^{K \times (X+1)}$ of each cluster:
\begin{equation}\label{eq:type_decoder}
Z_{c}^{type} = f_{\theta_{type}}[\tilde{x}_s \oplus \Phi_c \oplus \pi_c] \in \mathbb{R}^{K \times (X+1)},
\end{equation}
where $f_{\theta_{type}}$ is an FCN with learnable
parameters $\theta_{type}$. We consider $X = 5$ buffer types
and an additional ``None'' class, which indicates that no buffer
is inserted for the cluster.
Since the predicted buffer type affects the location decision,
its probability distribution is also provided to the location
decoder. The location decoder, \emph{Loc\_decoder},
predicts coordinates $Z_{c}^{loc} \in \mathbb{R}^{K \times 2}$
within the cluster bounding box
$[\mathbf{l}_c, \mathbf{u}_c]$, extended by 10\% of its extent
on each side:
\begin{equation}\label{eq:loc_decoder}
Z_{c}^{loc} = \mathbf{l}_c + \sigma\!\left(f_{\theta_{loc}}[\tilde{x}_s \oplus \Phi_c \oplus \mathrm{softmax}(Z_{c}^{type})]\right) \odot (\mathbf{u}_c - \mathbf{l}_c),
\end{equation}
where $f_{\theta_{loc}}$ is an FCN with learnable
parameters $\theta_{loc}$, and $\odot$ denotes element-wise
multiplication. Thus, the predicted buffer location is
constrained to the expanded bounding box of the cluster.
To decouple the tasks and improve robustness,
buffer locations are predicted for all clusters,
regardless of the buffer necessity
predicted by \emph{Type\_decoder}.
This approach mitigates potential error accumulation
caused by inaccurate buffer type predictions.

\noindent\textbf{Tree Construction: }
A buffer is generated for each non-empty cluster whose
predicted type is not ``None''. The predicted buffer location and input capacitance
are used to construct the node features for the next level,
together with the unbuffered nodes
(Section~\ref{subsec:training_strategy}).

\subsection{Training Strategy}
\label{subsec:training_strategy}

This section describes the training and inference
strategy for {\em MLBuf}.
We use buffer-embedded trees generated by OpenROAD 
Resizer (OR rsz) during global placement (\emph{repairDesign()} in RepairDesign.cc~\cite{rsz}) as 
the ground truth. 
We define the \emph{level} of a cell in the buffer-embedded tree
as the longest path from itself to the sink.
Suppose a buffer-embedded tree has $\tilde{H}$ levels (see
Figure~\ref{fig:learning_process}(c1));
it thus contains $\tilde{H}$ input-label pairs (see
Figure~\ref{fig:learning_process}(a2)(b2)(c2)) and can
be constructed through $\tilde{H}$ iterations.
More specifically, the input of the first iteration 
consists of the driver and all sinks. 
The corresponding label gives the sinks
``parent'' buffers in level
1 or None denoting no buffer. For the $i^{th} 
(1\leq i \leq \tilde{H})$ 
iteration, its input consists of the driver, the buffers 
predicted in the $(i-1)^{st}$ iteration, and the sinks whose 
parent buffer is None in the $(i-1)^{st}$ iteration.
Their respective labels are their parent
buffers in the level $i$ or None.
We also collect problematic nets for which OR rsz
does not insert any buffer as \emph{negative samples}.
Each negative sample contains a single input-label pair,
where the net-level gate is labeled ``no buffer''
and all sinks have a None buffer-parent label.
To balance the training data, we subsample negative
examples, limiting their number to 1.5 times the
number of buffered nets.

To mitigate error accumulation during training,
we adopt \emph{teacher forcing}~\cite{WilliamsZ89}.
Specifically, each training input is constructed from
the ground-truth buffers at the previous level, rather
than from the model's predictions.
This provides accurate supervision at each level.
For levels above the first, the driver's electrical
features (output capacitance and slew) are recomputed
using the feedback estimator described in
Appendix~\ref{subsec_app:model_details}.
The same estimator is used during inference to ensure
consistent electrical feature estimation.
The losses in Section~\ref{subsec:training_loss}
are computed independently for each level.
During inference, {\em MLBuf} recursively constructs
the buffer-embedded tree using predictions from the
previous level.
The training and inference procedures are provided
as Algorithms~\ref{alg:mlbuf_training} and
\ref{alg:mlbuf_inference}, respectively, in
Appendix~\ref{subsec_app:algo_mlbuf}.

\subsection{Training Loss}
\label{subsec:training_loss}

The training loss supervises the
net-level gate, cluster assignments, and predicted buffers.
Because the ordering of predicted clusters is arbitrary,
we match predicted clusters to ground-truth buffers one-to-one
before calculating the buffer-type and location losses.
This set-based formulation makes the loss independent of
cluster ordering.

\textbf{Gate Loss $L_g$:}
The gate $\hat{g}$ is trained with binary cross-entropy against the
label $g = 1$ if the ground-truth net contains at least one buffer
and $g = 0$ otherwise.
The loss is applied only at the first level,
because upper levels exist only for buffered nets.

\textbf{Cluster Loss $L_c$:}
We calculate co-membership probabilities using
$S = M_k M_k^T \in \mathbb{R}^{N \times N}$, where $S_{ij}$
is the probability that nodes $i$ and $j$ are assigned to
the same cluster. Binary cross-entropy is applied between
$S_{ij}$ and $y_{ij} \in \{0,1\}$, which indicates whether
the nodes share a parent buffer in the ground truth
(Appendix~\ref{subsec_app:model_details}). 

\textbf{Buffer Type Loss $L_{bt}$ and Location Loss $L_{bl}$:}
We use the Hungarian algorithm~\cite{Kuhn55}~\cite{CarionMSUKZ20}
to match predicted clusters
to ground-truth buffers one-to-one. The matching cost
accounts for the overlap between cluster members and the
children of a ground-truth buffer, as well as the distance
between predicted and ground-truth buffer locations.
For matched pairs, $L_{bt}$ is a class-weighted cross-entropy
loss over the $X+1$ buffer types, and $L_{bl}$ is a
smooth-$\ell_1$ location loss. Unmatched clusters are supervised using
type ``None'' through $L_{none}$.

\textbf{Count Loss $L_{cnt}$ and Chamfer Loss $L_{ch}$:}
The count loss $L_{cnt}$ penalizes the difference between
the expected number of predicted buffers
and the number of ground-truth buffers.
We further introduce a symmetric Chamfer loss~\cite{FanSG17}
$L_{ch}$ between predicted and ground-truth buffer locations,
weighted by the probabilities that predicted clusters
contain buffers. 
The total training loss is
$L = L_g + L_c + L_{bt} + L_{bl} + 0.2\,L_{none} + L_{cnt} + L_{ch}$.

\section{Experimental Validation}
\label{sec:experiments}

{\em MLBuf} is implemented in PyTorch.
It has been integrated with RePlAce on the
OpenROAD infrastructure to achieve
virtual buffering during global placement.
All codes and scripts are publicly released
in the Github repository~\cite{MLBuf}.
We run all experiments on a Linux server
with an AMD EPYC 7742 64-Core Processor CPU
(128 threads), 503GB RAM, and an NVIDIA 
A100-SXM4-80GB GPU. 
We evaluate {\em MLBuf} using eight NanGate45~\cite{NG45}
testcases from the OpenROAD-flow-scripts~\cite{ORFS}
and MacroPlacement~\cite{MacroPlacement} repositories.
Five testcases (ibex, jpeg, ariane, BlackParrot (BP) and
MegaBoom (MB)) were used in the MLCAD'25 version~\cite{KahngLW25}.
We add three testcases, swerv\_wrapper, tinyRocket and
dynamic\_node, in this revision.
Table~\ref{tab:testcases} summarizes the characteristics
of these benchmarks.
$\text{TCP}_{OR}$ (ns) and $\text{TCP}_{Invs}$ denote 
the target clock periods (TCP) used in the OpenROAD 
and Innovus\footnote{We do not perform any 
benchmarking of commercial EDA tools. Further, 
to avoid inadvertent benchmarking, we mask 
the target clock period values $\text{TCP}_{Invs}$
in Table~\ref{tab:testcases}.} flows, respectively. 

We use five sizes of buffers
from NanGate45, i.e., BUF\_X2, BUF\_X4,
BUF\_X8, BUF\_X16 and BUF\_X32.
We collect the training dataset using the OpenROAD
infrastructure. Specifically, we run the default
virtual buffering-based timing-driven global placer on
publicly available designs. Buffer insertion by OR rsz is
triggered when placement overflow reaches a threshold in
the predefined overflow\_list.\footnote{The default
 overflow\_list is \{0.7, 0.65, 0.6, 0.55, 0.5, 0.45,
 0.4, 0.35, 0.3, 0.25, 0.2, 0.15, 0.1\}; thus, the timing
 engine is invoked 13 times during global placement.}
At each timing optimization iteration, we collect
(i) the buffer-embedded trees constructed by OR rsz
(positive samples), and (ii) problematic nets examined
by OR rsz but left unbuffered (negative samples).
The dataset was collected from
six designs: ibex, jpeg, ariane133, ariane136, black\_parrot and bsg\_chip (BP).
For each design, we split nets into 85\% training and
15\% validation sets.
Hyperparameter selection and training details are
provided in Appendices~\ref{subsec_app:hyperparameter}
and~\ref{subsec_app:training_details}, respectively.
Table~\ref{tab:training_dataset} summarizes the
training dataset characteristics.
Four designs (swerv\_wrapper, tinyRocket, dynamic\_node and MB)
serve as unseen-design tests.
The collected dataset is publicly available at~\cite{MLBuf}.

\begin{table}[]

\caption{Characteristics of our benchmarks. }
\label{tab:testcases}
\resizebox{0.48\textwidth}{!}{%
\begin{tabular}{c|cccc}
\hline
Design (NG45)        & \multicolumn{1}{c}{\#Insts} & \multicolumn{1}{c}{\#Nets} & $\text{TCP}_{OR}$                 & \multicolumn{1}{c}{$\text{TCP}_{Invs}$} \\ \hline
ibex                 & 16907                        & 17728                       & 2.20 & -                        \\
jpeg                 & 53042                        & 58898                       & 1.40                  & -                        \\ 
ariane               & 119256                       & 142226                      & 4.00                  & -                        \\ 
BlackParrot (BP)     & 768851                       & 998716                      & NA                  & -                        \\ 
MegaBoom (MB)        & 1086920                      & 1443755                     & NA                    & -                        \\
swerv\_wrapper & 98169 & 105990 & 2.00 & NA \\
tinyRocket & 28330 & 32311 & 1.40 & NA \\
dynamic\_node & 11223 & 13585 & 5.50 & NA \\ \hline
\end{tabular}}
\end{table}

In this section, we first present 
the evaluation of 
predicted buffers in Section~\ref{subsec:comparison_with_baseline},
and the effect of {\em MLBuf-RePlAce} on the 
full placement and optimization flow in Section~\ref{subsec:ppa_val}.
We then provide the runtime breakdown of {\em MLBuf} in
Section~\ref{subsec:runtime}. Finally, we analyze 
the benefits of buffer-porosity-aware placement
with respect to end-to-end design quality in Section~\ref{subsec:case_study}.

\begin{table}[]
\caption{Characteristics of our training dataset.  
Buffered nets are counted per iteration (samples) and as distinct nets. Sinks and buffers per 
buffered net: [min, max].}
\label{tab:training_dataset}
\small
\centering
\resizebox{0.47\textwidth}{!}{%
\begin{tabular}{l|r|r|c|c|c}
\hline
Design  & \multicolumn{2}{c|}{Buffered nets} & Unbuffered  & Sinks & Buffers \\
           & samples & distinct &  nets & per net & per net \\ \hline
ibex &  1,773 & 158 & 57,329 & [2, 498] & [1, 17] \\
jpeg &  873 & 73 & 179,384 & [3, 2475] & [1, 54] \\
ariane133 & 34,800 & 2,965 & 77,276 & [1, 359] &  [1, 12] \\
ariane136 &  46,349 & 4,363 & 125,916 &  [1, 725] &  [1, 90] \\
black\_parrot &  9,127 & 1,142 & 46,306 &  [1, 1135] &  [1, 55] \\
bsg\_chip (BP) &  84,006 & 8,844 & 221,383 &  [1, 1758] &  [1, 33] \\  \hline
Total &  176,928 & 15,842 & 707,594 & [1, 2475] & [1, 90] \\ \hline
\end{tabular}}
\end{table}

\subsection{Evaluation of Predicted Buffer-embedded Trees}
~\label{subsec:comparison_with_baseline}
We compare the buffer-embedded trees predicted by
{\em MLBuf} against those generated by OR rsz. 
At each of the 13 timing optimization
iterations, we apply
{\em MLBuf} to the problematic nets and use the OR rsz
trees as the reference.
Table~\ref{tab:comparison_with_baseline} reports the
average buffer count and area over the 13 iterations.
We also report \emph{net recall}, the fraction of
OR rsz-buffered nets also buffered by {\em MLBuf};
\emph{net precision}, the fraction of {\em MLBuf}-buffered
nets also buffered by OR rsz; and \emph{size accuracy},
the fraction of matched buffers with the same type
as the OR rsz reference.
{\em MLBuf} buffers all nets buffered by OR rsz and
some additional nets, with net precisions of 0.69
and 0.71 for ibex and jpeg, respectively.
It predicts a comparable number of buffers while reducing
the total buffer area by 8\% and 23\%, respectively.
The predicted buffer types match OR rsz in 92\% and
86\% of matched buffers. 

\begin{table}[]
\caption{Evaluation of predicted buffer-embedded trees
(averages over the 13 timing optimization iterations).}
\label{tab:comparison_with_baseline}
\footnotesize
\centering
\resizebox{0.47\textwidth}{!}{%
\begin{tabular}{|c|c|c|c|c|c|c|}
\hline
Design & Method & \#Buffers & Buffer area ($\mu m^2$) & Net recall & Net precision & Size accuracy \\ \hline
\multirow{2}{*}{ibex} & OR rsz & 136 & 649  & -    & -    & -    \\ \cline{2-7}
                      & MLBuf  & 151 & 596  & 1.00 & 0.69 & 0.92 \\ \hline
\multirow{2}{*}{jpeg} & OR rsz & 158 & 1229 & -    & -    & -    \\ \cline{2-7}
                      & MLBuf  & 129 & 944  & 1.00 & 0.71 & 0.86 \\ \hline
\end{tabular}}
\end{table}

\needspace{5\baselineskip}
\subsection{PPA Validation}~\label{subsec:ppa_val}

We now present post-route PPA results 
of {\em MLBuf-RePlAce} obtained using both the open-source 
OpenROAD flow \cite{AjayiCFHH2019} and the commercial flow.
Since no existing ML-driven buffering method is integrated
into analytical placement to complete the full ``closed-loop''
optimization flow, we evaluate the effectiveness of 
{\em MLBuf-RePlAce} (runscripts are in the Github 
repository~\cite{MLBuf}) by comparing it against the 
following three approaches:

\begin{itemize}[noitemsep, topsep=0pt, leftmargin=*]
\item RePlAce: Pure RePlAce without timing-driven mode. This
is the state-of-the-art open-source global placer in the 
OpenROAD project~\cite{OpenROAD} (commit hash: df581be).

\item TD-RePlAce: Default virtual buffering-based 
timing-driven global placement in OpenROAD.
In this mode, OR rsz~\cite{rsz} is 
invoked during global placement to repair 
nets with ERC violations by inserting buffers.
It updates net weights based on slack estimation
and subsequently removes buffers to achieve 
virtual buffering (commit hash: df581be).

\item Ad-Hoc Baseline: This is a rule-based analytical 
approximation used as a baseline for comparison. 
Detailed description of this approach is provided
in Appendix~\ref{subsec_app:ad_hoc}.
Note that all hyperparameters in this approach are 
tuned to enhance the accuracy of estimated buffer 
counts by benchmarking against 
OR rsz, thereby improving
the competitiveness of this baseline.

\end{itemize}

We use the values of post-route wirelength, WNS, TNS,
and power (with respect to TCP) for purposes of comparison.
We repeat each configuration four times.
For RePlAce, TD-RePlAce and
{\em MLBuf-RePlAce}, run 0 uses the original placement.
For runs 1--3, we perturb the run-0 placement before the
downstream flow by adding independent Gaussian noise
$N(0, 0.01\,\mu m)$ to both coordinates of each standard cell
(approximately 1/20 of a NanGate45 placement site).
This measures the sensitivity of downstream optimization
to small placement perturbations. Ad-Hoc runs use different
random seeds.
Tables~\ref{tab:post_route_openroad}
and~\ref{tab:post_route_invs} report the mean and standard
deviation of the results.

\noindent\textbf{PPA Validation with OpenROAD Flow. }
The flow begins with synthesis using Yosys, 
and floorplanning using OpenROAD.
The synthesized netlist is debuffered, and 
then global placement is performed using RePlAce. 
During global placement, 
timing optimization is triggered
when the placement overflow reaches each threshold 
specified in the predefined 
overflow\_list.
At each optimization iteration, problematic nets are 
identified through OpenSTA and fed into one of three
virtual buffering strategies, 
{\em MLBuf}, OR rsz, or the Ad-Hoc baseline.
When {\em MLBuf} or the Ad-Hoc baseline 
is employed, the bin densities are updated 
according to the predicted 
buffer types and locations (see Section~\ref{sec:approach}), 
enabling the placer to account for buffer area effects.
When OR rsz is employed, we use the 
default virtual buffering strategy in OpenROAD, 
which adjusts net weights based on the estimated slacks.
Subsequent steps correspond to OpenROAD's standard flow:
resizing, buffering, legalization, detailed
placement, clock tree synthesis (CTS), 
and routing. We compare the post-route PPA results
in Table~\ref{tab:post_route_openroad}.
We exclude ariane from the comparison, as all of its data collected from OpenROAD is used in training.
BP and MB are excluded because the
OpenROAD flow fails to route these designs.
Table~\ref{tab:post_route_openroad} shows that,
compared with TD-RePlAce, {\em MLBuf-RePlAce} obtains
the best TNS on four of the five designs.
It achieves (maximum, average) improvements of
(49\%, 24\%) in TNS and (38\%, 19\%) in WNS,
with post-route power within 1\% of TD-RePlAce.

\begin{table}[]
\caption{Post-route metrics evaluated with OpenROAD flow (mean $\pm$ standard deviation over four runs). rWL in $\mu m$, WNS and TNS in ns, and power in mW.}
\label{tab:post_route_openroad}
\centering
\resizebox{0.48\textwidth}{!}{%
\begin{tabular}{|c|c|c|c|c|c|}
\hline
Design  & Method & rWL & WNS & TNS & Power \\ \hline
\multirow{4}{*}{ibex } & RePlAce & 314160 & -0.287 $\pm$ 0.030 & -29.4 $\pm$ 4.9 & 131.5 \\ \cline{2-6}
 & TD-RePlAce & 311973 & -0.276 $\pm$ 0.022 & -29.0 $\pm$ 4.0 & 130.2 \\ \cline{2-6}
 & Ad-Hoc & 302604 & -0.244 $\pm$ 0.025 & -19.0 $\pm$ 3.2 & 128.9 \\ \cline{2-6}
 & MLBuf & 302952 & \textbf{-0.224 $\pm$ 0.013} & \textbf{-15.3 $\pm$ 3.4} & 128.9 \\ \hline\hline
\multirow{4}{*}{jpeg} & RePlAce & 634668 & \textbf{-0.100 $\pm$ 0.014} & -5.0 $\pm$ 1.3 & 561.7 \\ \cline{2-6}
 & TD-RePlAce & 635628 & -0.139 $\pm$ 0.022 & -4.8 $\pm$ 0.6 & 556.8 \\ \cline{2-6}
 & Ad-Hoc & 632978 & -0.113 $\pm$ 0.011 & -5.8 $\pm$ 2.2 & 554.6 \\ \cline{2-6}
 & MLBuf & 631643 & -0.104 $\pm$ 0.010 & \textbf{-4.3 $\pm$ 0.7} & 551.2 \\ \hline\hline
\multirow{4}{*}{swerv\_wrapper } & RePlAce & 3901282 & -0.800 $\pm$ 0.024 & -551.4 $\pm$ 9.8 & 290.0 \\ \cline{2-6}
 & TD-RePlAce & 3920689 & -0.684 $\pm$ 0.062 & -493.1 $\pm$ 92.6 & 289.2 \\ \cline{2-6}
 & Ad-Hoc & 3921169 & -0.619 $\pm$ 0.035 & -472.9 $\pm$ 25.6 & 289.6 \\ \cline{2-6}
 & MLBuf & 3932026 & \textbf{-0.604 $\pm$ 0.037} & \textbf{-410.1 $\pm$ 34.9} & 289.9 \\ \hline\hline
\multirow{4}{*}{tinyRocket } & RePlAce & 630172 & -0.419 $\pm$ 0.031 & -54.3 $\pm$ 11.8 & 128.7 \\ \cline{2-6}
 & TD-RePlAce & 629616 & -0.359 $\pm$ 0.049 & -45.0 $\pm$ 12.9 & 128.8 \\ \cline{2-6}
 & Ad-Hoc & 629592 & -0.277 $\pm$ 0.049 & -30.2 $\pm$ 6.2 & 129.1 \\ \cline{2-6}
 & MLBuf & 632830 & \textbf{-0.221 $\pm$ 0.020} & \textbf{-23.0 $\pm$ 5.4} & 129.6 \\ \hline\hline
\multirow{4}{*}{dynamic\_node } & RePlAce & 204192 & -0.686 $\pm$ 0.008 & -149.8 $\pm$ 4.4 & 18.5 \\ \cline{2-6}
 & TD-RePlAce & 206757 & -0.707 $\pm$ 0.035 & \textbf{-144.0 $\pm$ 0.6} & 18.8 \\ \cline{2-6}
 & Ad-Hoc & 206040 & \textbf{-0.667 $\pm$ 0.006} & -148.3 $\pm$ 3.7 & 18.8 \\ \cline{2-6}
 & MLBuf & 204796 & -0.689 $\pm$ 0.011 & -149.0 $\pm$ 2.1 & 18.7 \\ \hline
\end{tabular}}
\end{table}

\noindent\textbf{PPA Validation with Commercial Flow. }
The flow begins with synthesis using Cadence
Genus 21.1, a state-of-the-art commercial synthesis 
tool. The synthesized netlist is then debuffered, 
and floorplanning is conducted using Cadence 
Innovus 21.1. Global placement is subsequently 
performed using RePlAce within the OpenROAD framework,
following the same methodology as described in 
the OpenROAD-based flow. During global placement, 
{\em MLBuf}, OR rsz or the Ad-Hoc baseline 
is invoked to perform virtual buffering.
After global placement, the
default optimization flow provided by Innovus 
is executed to complete the physical design
and produce post-route PPA results.
We compare post-route wirelength, WNS, TNS,
power and number of failing endpoints
(\#FEP) in Table~\ref{tab:post_route_invs}.
{\em MLBuf-RePlAce}
achieves the best mean TNS on all five designs and
the fewest failing endpoints on four.
Compared with TD-RePlAce, it achieves (maximum, average)
improvements of (56\%, 51\%) in WNS, (83\%, 69\%) in TNS,
and (60\%, 38\%) in \#FEP.
Section~\ref{subsec:case_study} further examines the
impact of the clock period on these results.

\begin{table}[]
\caption{Post-route metrics evaluated with Commercial flow. rWL in $\mu m$, WNS and TNS in ns, and power in mW.}
\label{tab:post_route_invs}
\centering
\resizebox{0.48\textwidth}{!}{%
\begin{tabular}{|c|c|c|c|c|c|c|}
\hline
Design & Method & rWL & WNS & TNS & Power & \#FEP \\ \hline
\multirow{4}{*}{ibex } & RePlAce & 290072 & -0.058 & -28.8 $\pm$ 11.9 & 44.3 & 1162 \\ \cline{2-7}
 & TD-RePlAce & 278069 & -0.081 & -57.2 $\pm$ 20.9 & 42.7 & 1363 \\ \cline{2-7}
 & Ad-Hoc & 279729 & -0.071 & -31.2 $\pm$ 0.9 & 43.2 & 1094 \\ \cline{2-7}
 & MLBuf & 290992 & \textbf{-0.053} & \textbf{-21.4 $\pm$ 0.4} & 44.7 & 1099 \\ \hline\hline
\multirow{4}{*}{jpeg } & RePlAce & 581964 & -0.061 & -23.5 $\pm$ 18.0 & 487.2 & 1202 \\ \cline{2-7}
 & TD-RePlAce & 573553 & -0.101 & -50.4 $\pm$ 9.4 & 479.3 & 1508 \\ \cline{2-7}
 & Ad-Hoc & 586305 & -0.067 & -24.5 $\pm$ 17.4 & 478.4 & 1191 \\ \cline{2-7}
 & MLBuf & 574834 & \textbf{-0.044} & \textbf{-16.6 $\pm$ 8.9} & 482.0 & 1148 \\ \hline\hline
\multirow{4}{*}{ariane } & RePlAce & 4745347 & -0.074 & -30.4 $\pm$ 23.2 & 793.4 & 1219 \\ \cline{2-7}
 & TD-RePlAce & 4762765 & -0.062 & -22.6 $\pm$ 23.0 & 794.6 & 912 \\ \cline{2-7}
 & Ad-Hoc & 4757820 & -0.054 & -21.4 $\pm$ 1.5 & 792.1 & 1122 \\ \cline{2-7}
 & MLBuf & 4772096 & \textbf{-0.027} & \textbf{-3.8 $\pm$ 2.1} & 793.3 & 455 \\ \hline\hline
\multirow{4}{*}{BP } & RePlAce & 27186523 & -0.034 & -11.2 $\pm$ 2.8 & 3991.5 & 1568 \\ \cline{2-7}
 & TD-RePlAce & 27144580 & -0.032 & -8.7 $\pm$ 7.6 & 3989.6 & 1166 \\ \cline{2-7}
 & Ad-Hoc & 27046677 & -0.025 & -9.8 $\pm$ 6.8 & 3988.2 & 1421 \\ \cline{2-7}
 & MLBuf & 27208718 & \textbf{-0.015} & \textbf{-3.3 $\pm$ 3.3} & 3988.8 & 732 \\ \hline\hline
\multirow{4}{*}{MB } & RePlAce & 41212741 & -0.049 & -2.9 $\pm$ 3.0 & 2153.3 & 126 \\ \cline{2-7}
 & TD-RePlAce & 40411910 & -0.071 & -3.4 $\pm$ 2.2 & 2159.6 & 122 \\ \cline{2-7}
 & Ad-Hoc & 40602288 & -0.068 & -3.1 $\pm$ 1.3 & 2164.1 & 103 \\ \cline{2-7}
 & MLBuf & 39830881 & \textbf{-0.032} & \textbf{-1.0 $\pm$ 1.1} & 2136.2 & 49 \\ \hline
\end{tabular}}
\end{table}

\subsection{Runtime Breakdown}~\label{subsec:runtime}
\begin{figure}[htbp]
\includegraphics[scale=0.22]{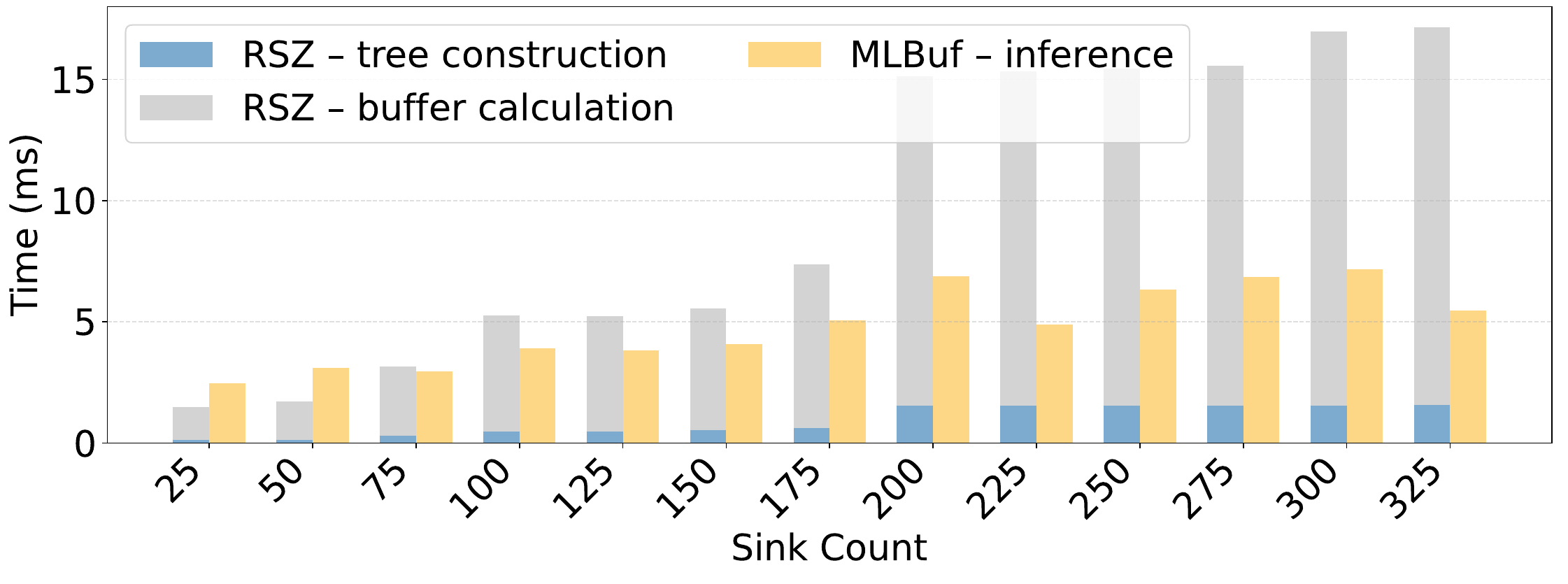}
\centering  
\caption{Runtime comparison between MLBuf and OR rsz}.\label{fig:runtime}
\end{figure}

Figure~\ref{fig:runtime} provides a runtime comparison
between OR rsz and {\em MLBuf}. OR rsz runs 
on a single CPU thread. The runtime contains the Steiner tree
generation and buffer calculation (i.e., buffer type
selection and buffer location calculation). 
For {\em MLBuf}, we assume that the features of the driver
and sinks are already extracted, and we measure the runtime
of generating an entire buffer-embedded tree
(i.e., inference time). As net sizes increase, 
the runtime of OR rsz increases significantly. 
On the other hand, {\em MLBuf}
consistently predicts buffer types and location with high
efficiency, achieving over a 3$\times$ speedup on large nets compared
to OR rsz. These results demonstrate the efficiency
and scalability of {\em MLBuf}.
Since each net can be considered independently during virtual
buffering, our ongoing work further enhances the efficiency
of {\em MLBuf-RePlAce} by parallelizing 
net processing.

\subsection{Benefits of Buffer-porosity-aware Placement}~\label{subsec:case_study}
This section analyzes the impact of 
buffer-porosity-aware placement on the 
full optimization flow, particularly under
varying timing constraints. 
We systematically sweep the target clock period (TCP),
from 0.98ns to 1.35ns,
to adjust the stringency of timing requirements. 
Decreasing the TCP tightens these timing constraints, 
and hence requires more buffer insertions for timing optimization. 


\begin{figure}[htbp]
\includegraphics[scale=0.4]{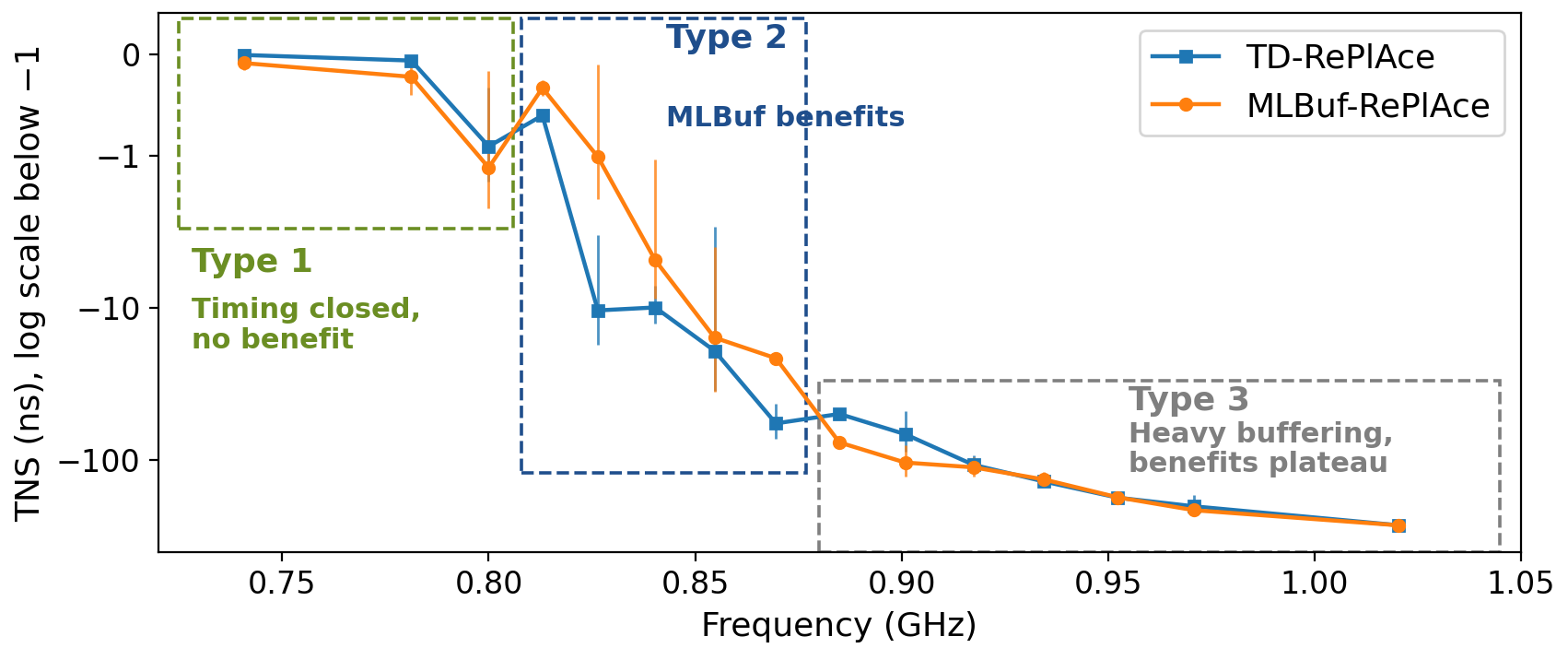}
\centering  
\caption{Post-route TNS comparisons between {\em MLBuf-RePlAce} 
  and TD-RePlAce under different clock frequencies.}\label{fig:tcp}
\end{figure}

Figure~\ref{fig:tcp} shows the post-route TNS
of {\em MLBuf-RePlAce} and TD-RePlAce across different clock frequencies ($\mathrm{GHz}$) ($clock\_frequency  = \frac{1}{TCP}$) on 
ibex under the OpenROAD infrastructure. 
We observe three {\em Types} of outcomes: 
(i) Type 1, where there is no buffering; 
(ii) Type 2, where {\em MLBuf} consistently outperforms TD-RePlAce; 
and (iii) Type 3, where benefits plateau with heavy buffering.
In general, {\em MLBuf-RePlAce} delivers greater 
benefits with greater buffering needs, i.e., with Type 2 outcomes. 
{\em MLBuf} guides the placer to pre-allocate buffer space for  
downstream optimization, thereby reducing
routing detours and achieving better PPA results. 
Our results also confirm the importance of buffer-porosity-aware 
global placer for the full optimization flow.
However, as {\em MLBuf} is designed for resolving ERC 
violations and is insensitive to timing constraints, the 
improvements plateau when the TCP becomes too tight 
(Type 3), indicating that 
the number of buffers predicted by {\em MLBuf} is 
insufficient under extremely constrained timing. 
Our ongoing work further enhances {\em MLBuf} by incorporating 
timing-related features and constraints, enabling timing-aware
virtual buffering prediction.

\section{Conclusion and Future Directions}
\label{sec:conclusion}

We have presented {\em MLbuf} and {\em MLBuf-RePlAce}, 
a learning-driven virtual buffering-aware 
analytical global placement framework 
built upon the OpenROAD infrastructure.
Experimental results demonstrate that
{\em MLBuf-RePlAce} outperforms the
default virtual buffering-based global placer
in OpenROAD in terms of post-route WNS and TNS
metrics,
with no post-route power degradation.
Ongoing extensions to {\em MLBuf-RePlAce} include:
(i) enabling inverter-based repeater insertion and handling sink polarity,
(ii) incorporating delay information into the virtual 
buffering model to enable net-weighted timing-driven 
global placement, and (iii) integrating the virtual 
buffering model with state-of-the-art GPU-accelerated 
global placers \cite{KahngW25}\cite{LiaoLCLLY22}
to support virtual buffering-aware timing-driven 
GPU-accelerated global placement.


\section*{Acknowledgments}
This work is partially supported by the Samsung AI Center.
We sincerely thank Cezar Trisca Vicol and his colleagues at Partcl for identifying and reporting the issue~\cite{Issue6}
in our previous version.

\vspace{12pt}

\newpage
\clearpage

\appendices
\section{Supplementary Appendix}

\subsection {MLBuf Model Details}
~\label{subsec_app:model_details}

\textbf{Input Feature List for MLBuf:}
The set encoder uses the following features
for each node (Section~\ref{subsec:model_structure}):
\begin{itemize}
\item A three-dimensional one-hot encoding of the node
type (driver, sink or buffer);
\item Node coordinates relative to the driver
(placed at (0, 0));
\item Manhattan distance from the node to the driver;
\item Input capacitance (fF) and input
slew (ps) for sinks and buffers (zero for the driver);
\item Driver output slew (ps), output load
capacitance (fF), maximum capacitance (fF) and output
resistance ($\Omega$) (zero for sinks and buffers); and
\item Ratio of the driver's output capacitance to its
maximum capacitance, rescaled
to $[0, 1]$ (zero for sinks and buffers).
\end{itemize}
The net-context features are:
(i) the number of nodes, net HPWL,
total sink capacitance and estimated load
(Eq.~\ref{eq:output_cap_update});
(ii) the ratio of driver load to maximum capacitance,
and an indicator of whether this ratio exceeds 1;
(iii) recursion level normalized by the maximum depth (5);
and (iv) the driver's output slew.

\textbf{Feature Update Formulas During Inference:}
As described in Section~\ref{subsec:training_strategy},
MLBuf dynamically updates the output load capacitance and slew of
cells during model inference.

The output load capacitance of the driver is estimated as
the sum of the wire capacitance $C(wire)$ and the input
pin capacitances of its child cells $C_{in}(p_{child})$:
\begin{equation}~\label{eq:output_cap_update}
\begin{aligned}
& C_{out}(p_{drvr}) = a \cdot C(wire)+ \sum{C_{in}(p_{child})},\\
& \quad C(wire) = \alpha \mathcal{W} \times C_{wire\_unit},
\end{aligned}
\end{equation}
where $\mathcal{W}$ is the half-perimeter wirelength (HPWL)
of the bounding box enclosing the driver and its children,
$\alpha$ $= 1.2$ is a scaling factor,
$C_{wire\_unit}$ is the unit wire capacitance (0.0889\,fF/$\mu m$ in NanGate45), and $a$ is a fitted coefficient.

The output slew of the driver is estimated as
\begin{equation}~\label{eq:output_slew_update}
\begin{aligned}
& S_{out}(p_{drvr}) = b\, R_{drvr}\, C_{out}(p_{drvr}) + c\, \mathcal{W}\, C_{out}(p_{drvr}) + d,
\end{aligned}
\end{equation}
where $R_{drvr}$ is the driver’s output resistance, and $b$, $c$ and $d$ are fitting coefficients
that capture wire resistance and higher-order delay effects.
With $R_{drvr}$ in k$\Omega$, $C_{out}$ in fF,
$\mathcal{W}$ in mm and $S_{out}$ in ps, the released
model uses $a = 0.169$, $b = 0.603$, $c = 0.0488$
and $d = 75.6$. These coefficients are fitted by
least squares to OpenSTA values from intermediate
snapshots of the Resizer-generated trees, after the
driver's children are replaced by buffers.
The same estimator is used to generate upper-level
input features during training. The input slew of
each child cell connected to the driver is estimated
to be equal to the driver's output slew.

\textbf{Cluster Loss Formulation:}
As described in Section~\ref{subsec:training_loss},
$L_c$ measures cluster co-membership.
Let $S = M_k M_k^T$, where $M_k$ is the soft assignment
matrix of Eq.~(\ref{eq:gumbel_softmax}) without noise.
Then, $S_{ij}$ represents the probability that nodes
$i$ and $j$ belong to the same cluster. The loss is
\begin{equation}
\begin{aligned}
L_{c} &= \tfrac{1}{2}\Big( \langle \ell_{ij} \rangle_{y_{ij}=1} + \langle \ell_{ij} \rangle_{y_{ij}=0} \Big), \\
\ell_{ij} &= -y_{ij} \log S_{ij} - (1-y_{ij}) \log (1 - S_{ij}),
\end{aligned}
\end{equation}
where $y_{ij} \in \{0, 1\}$ indicates whether
nodes $i$ and $j$ share a parent buffer in the ground
truth. The operator $\langle \cdot \rangle$ averages over
node pairs for which at least one node is buffered.
Positive pairs encourage similar
cluster assignments, whereas negative pairs encourage
different assignments.

\textbf{Matching, Count and Chamfer Loss Formulations:}
Let $\{(t_q, \mathbf{z}_q)\}_{q=1}^{Q}$ denote
the buffer types and locations of the $Q$ ground-truth
buffers at one level, and let $B_q$ denote the nodes
driven by buffer $q$. For cluster $k$, let
$p_k = \mathrm{softmax}(Z^{type}_{c})_k$ be the predicted
type distribution, and let $Z^{loc}_{c,k}$ be the
predicted location. The matching cost is
\begin{equation}
\begin{aligned}
\mathrm{cost}(k, q) &= -\frac{1}{|B_q|}\sum_{i \in B_q} [M_k]_{ik} + \frac{\|Z^{loc}_{c,k} - \mathbf{z}_q\|_1}{s}, \\
s &= \mathcal{W}/4 + 5\,\mu m,
\end{aligned}
\end{equation}
The Hungarian algorithm determines the minimum-cost
one-to-one matching $\mathcal{M}$.
For each matched pair $(k, q) \in \mathcal{M}$, we compute
\begin{equation}
L_{bt} = \mathrm{CE}_w\big(Z^{type}_{c,k}, t_q\big), \qquad
L_{bl} = \mathrm{smooth}\text{-}\ell_1\!\left(\frac{Z^{loc}_{c,k}}{s}, \frac{\mathbf{z}_q}{s}\right),
\end{equation}
where $\mathrm{CE}_w$ denotes cross-entropy with
class weights inversely proportional to buffer-type
frequency. Unmatched clusters contribute
$L_{none} = \mathrm{CE}_w(Z^{type}_{c,k}, \mathrm{None})$.
Let $b_k = 1 - p_{k, \mathrm{None}}$ denote the predicted
probability that cluster $k$ contains a buffer, and
$d_{kq} = \|Z^{loc}_{c,k} - \mathbf{z}_q\|_1 / s$
denote the normalized location distance.
\begin{equation}
\begin{aligned}
L_{cnt} &= \frac{\mathrm{smooth}\text{-}\ell_1\big(\textstyle\sum_k b_k,\; Q\big)}{\max(Q, 1)}, \\
L_{ch} &= \frac{1}{Q}\sum_{q} \min_k \big( d_{kq} + 10\,(1 - b_k) \big)
        + \frac{\sum_k b_k \min_q d_{kq}}{\sum_k b_k}.
\end{aligned}
\end{equation}
The first term of $L_{ch}$ encourages a
high-confidence prediction near every ground-truth
buffer. The second term penalizes predicted buffers
that are far from all ground-truth buffers.
The count and Chamfer losses, together
with the matched losses, replace the ERC, wirelength
and area penalties of the MLCAD'25 version~\cite{KahngLW25}.

\subsection{Algorithm Flows of MLBuf Training and Inference}
\label{subsec_app:algo_mlbuf}

Algorithm~\ref{alg:mlbuf_training} and Algorithm~\ref{alg:mlbuf_inference}
respectively describe {\em MLBuf}'s training and inference
procedures. See the discussion in Section~\ref{subsec:training_strategy}.

\textbf{{\em MLBuf} training flow (Algorithm~\ref{alg:mlbuf_training}).}
Each ground-truth buffer-embedded tree
$T \in \mathcal{D}_{train}$ with $\tilde{H}$ levels
is converted into $\tilde{H}$ input-label pairs
$(\{\mathbf{X}_i\}_{i=1}^{\tilde{H}},
\{\mathcal{Y}_i\}_{i=1}^{\tilde{H}})$.
An unbuffered (negative) net yields one pair with an
empty label (Lines 1--2).
During training, we apply teacher forcing (Lines 4--15).
At iteration $i \in [1, \tilde{H}]$, {\em MLBuf}
predicts the gate $\hat{g}_i$, the cluster assignment
matrix $\mathbf{M}_i$, buffer types $\mathbf{Z}^{type}_i$
and buffer locations $\mathbf{Z}^{loc}_i$ (Lines 5--7).
We then calculate the gate and co-membership losses
(Lines 8--9), match clusters to ground-truth buffers using
the Hungarian algorithm, and calculate the matched type
and location losses (Lines 10--11), followed by the count
and Chamfer losses (Line 12).
The next iteration uses the $(i+1)^{st}$ ground-truth
input-label pair (Line 14). After all $\tilde{H}$
iterations, the model parameters are updated by
back-propagation (Line 16).
In implementation, levels of multiple nets are processed
together using mini-batches grouped by sequence length.

\begin{algorithm}[htp]
\renewcommand{\algorithmicrequire}{\textbf{Input:}}
\renewcommand{\algorithmicensure}{\textbf{Output:}}
\caption{MLBuf Training Procedure}
\label{alg:mlbuf_training}
\begin{algorithmic}[1]
\REQUIRE Training set $\mathcal{D}_{\text{train}}=\{T_1,\dots,T_N\}$ of buffered (positive) and unbuffered (negative) nets, maximum tree depth $\tilde{H}$, learnable parameters $\delta$, learning rate $\eta$
\ENSURE Optimized model parameters $\delta^\star$
\FOR{\textbf{each} net $T \in \mathcal{D}_{\text{train}}$}
    \STATE $(\{\mathbf{X}_i\}_{i=1}^{\tilde{H}},\{\mathcal{Y}_i\}_{i=1}^{\tilde{H}})\leftarrow\textsc{BuildPairs}(T)$  \hfill $\triangleright$ \COMMENT{$\tilde{H}$ input-label pairs; upper-level features
from the feedback estimator; $\tilde{H}=1$ and
$\mathcal{Y}_1=\emptyset$ for a negative net}
    \STATE $L_{\text{total}}\leftarrow 0$
    \FOR{$i=1$ \TO $\tilde{H}$}
        \STATE $(\hat{g}_i, \mathbf{M}_i)\leftarrow\mathcal{F}_{c}(\mathbf{X}_i;\delta)$
        \hfill $\triangleright$ \COMMENT{gate and} cluster assignment matrix
        \STATE $\mathbf{Z}^{\text{type}}_i\leftarrow\mathcal{F}_{\text{type}}(\mathbf{X}_i,\mathbf{M}_i;\delta)$ \hfill $\triangleright$ \COMMENT{buffer types}
        \STATE $\mathbf{Z}^{\text{loc}}_i\leftarrow\mathcal{F}_{\text{loc}}(\mathbf{X}_i,\mathbf{M}_i,\mathbf{Z}^{\text{type}}_i;\delta)$ \hfill $\triangleright$ \COMMENT{buffer locations}
        \STATE $L_g \leftarrow \textsc{GateLoss}(\hat{g}_i,\mathcal{Y}_i)$ \textbf{if} $i=1$ \textbf{else} $0$
        \STATE $L_c \leftarrow \textsc{CoMembershipLoss}(\mathbf{M}_i,\mathcal{Y}_i)$
        \STATE $\mathcal{M}\leftarrow\textsc{Hungarian}(\mathbf{M}_i,\mathbf{Z}^{\text{loc}}_i,\mathcal{Y}_i)$ \hfill $\triangleright$ \COMMENT{clusters $\leftrightarrow$ ground-truth buffers}
        \STATE $(L_{bt},L_{bl},L_{none})\leftarrow\textsc{MatchedLosses}(\mathbf{Z}^{\text{type}}_i,\mathbf{Z}^{\text{loc}}_i,\mathcal{M})$
        \STATE $(L_{cnt},L_{ch})\leftarrow\textsc{CountChamfer}(\mathbf{Z}^{\text{type}}_i,\mathbf{Z}^{\text{loc}}_i,\mathcal{Y}_i)$
        \STATE $L_{\text{total}}\mathrel{+}= L_g + L_c + L_{bt} + L_{bl} + 0.2\,L_{none} + L_{cnt} + L_{ch}$
        \STATE \COMMENT{\em the \textit{(i+1)}‑th ground‑truth pair is used automatically via $\mathbf{X}_{i+1},\mathcal{Y}_{i+1}$}
    \ENDFOR
    \STATE $\delta\leftarrow\delta-\eta\,\nabla_\delta L_{\text{total}}$ \hfill $\triangleright$ \COMMENT{back‑propagation}
\ENDFOR
\end{algorithmic}
\end{algorithm}

\textbf{{\em MLBuf} inference flow (Algorithm~\ref{alg:mlbuf_inference}).}
Given the driver and sinks of a problematic net
(Line 3), the pretrained model $\mathcal{F}_{\delta^*}$
predicts the net-level gate and hard cluster assignments
(Line 4). If the gate probability is below the threshold,
the net is left unbuffered (Lines 5--7).
Otherwise, the model predicts the buffer types and
locations (Lines 8--9) and inserts a buffer for every
non-empty cluster whose type is not None (Line 10).
Inference terminates when no buffer is predicted (Lines 2, 11--13).
At each level, we update the electrical features of
predicted buffers and remaining unbuffered sinks using
Eqs.~(\ref{eq:output_cap_update}) and
(\ref{eq:output_slew_update}) (Line 14).
These features form the input to the next iteration
(Lines 15--17). The final output is the complete
buffer-embedded tree (Line 19).

\begin{algorithm}[htp]
\renewcommand{\algorithmicrequire}{\textbf{Input:}}
\renewcommand{\algorithmicensure}{\textbf{Output:}}
\caption{MLBuf Inference Procedure}
\label{alg:mlbuf_inference}
\begin{algorithmic}[1]
\REQUIRE Driver $d$ and initial sink set $\mathcal{V}^{s}_{0}$, trained model $\mathcal{F}_{\delta^\star}$, gate threshold $g_{thr}$, maximum depth $H_{max}$
\ENSURE Predicted buffer–embedded tree $\hat{T}$
\STATE Initialize index $i\leftarrow 1$, buffer set $\mathcal{V}^{b}_{0}\leftarrow\emptyset$, tree $\hat{T}\leftarrow\emptyset$
\WHILE{$i \le H_{max}$}
    \STATE $\mathbf{X}_i\leftarrow\{\mathcal{V}^{b}_{i-1},\,\mathcal{V}^{s}_{i-1}\}$
    \hfill $\triangleright$ \COMMENT{features of current buffers+sinks}
    \STATE $(\hat{g}_i,\mathbf{M}_i)\leftarrow\mathcal{F}_{c}(\mathbf{X}_i;\delta^\star)$ \hfill $\triangleright$ \COMMENT{gate; arg-max} cluster assignment
    \IF{$i = 1$ \textbf{and} $\hat{g}_1 < g_{thr}$}
        \STATE \textbf{break} \hfill $\triangleright$ \COMMENT{no buffering is required}
    \ENDIF
    \STATE $\mathbf{Z}^{\text{type}}_i\leftarrow\mathcal{F}_{\text{type}}(\mathbf{X}_i,\mathbf{M}_i;\delta^\star)$ \hfill $\triangleright$ \COMMENT{buffer types}
    \STATE $\mathbf{Z}^{\text{loc}}_i\leftarrow\mathcal{F}_{\text{loc}}(\mathbf{X}_i,\mathbf{M}_i,\mathbf{Z}^{\text{type}}_i;\delta^\star)$ \hfill $\triangleright$ \COMMENT{buffer locations}
    \STATE $\mathcal{V}^{b}_{i}\leftarrow\textsc{ExtractBuffers}(\mathbf{M}_i,\,\mathbf{Z}^{\text{type}}_i,\mathbf{Z}^{\text{loc}}_i)$ \hfill $\triangleright$ \COMMENT{non-empty clusters with type $\neq$ \textsc{None}}
    \IF{$\mathcal{V}^{b}_{i}=\emptyset$ }
        \STATE \textbf{break}
    \ENDIF
    \STATE Update electrical features of $\mathcal{V}^{b}_{i}$ and $\mathcal{V}^{s}_{i-1}$ using Eq.~\eqref{eq:output_cap_update} and Eq.~\eqref{eq:output_slew_update}
    \STATE $\hat{T}\leftarrow\hat{T}\cup\mathcal{V}^{b}_{i}$ \hfill
 $\triangleright$ \COMMENT{merge new buffers into the tree}
    \STATE $\mathcal{V}^{s}_{i}\leftarrow$ unbuffered sinks after inserting $\mathcal{V}^{b}_{i}$
    \STATE $i\leftarrow i+1$
\ENDWHILE
\RETURN $\hat{T}$
\end{algorithmic}
\end{algorithm}

\subsection{Experimental Details}
\label{subsec_app:experimental_details}

\subsubsection{Hyperparameter Selection}
\label{subsec_app:hyperparameter}
The main hyperparameters of {\em MLBuf} are the
number of clusters $K$ and the net-level gate threshold.
$K$, the output dimension of the clustering module,
limits the number of buffers predicted at each level.
We set $K = 32$ and no validation
level reaches this limit.
The gate threshold is selected to maximize the F1
score of the buffered-versus-unbuffered classification
on the validation set. The selected value is 0.40,
with an F1 score of 0.985.
We anneal the Gumbel-Softmax temperature from 1.0
to 0.3 during the first 12 epochs and do not use
straight-through hard assignments during training.
Other settings follow common transformer practice
and are listed in Table~\ref{tab:MLBuf_hyperparameter}.


\subsubsection{MLBuf Training Details}~\label{subsec_app:training_details}
We collect training data from OpenROAD using the
procedure described in Section~\ref{sec:experiments}.
For each design, 85\% of nets are used for training
and 15\% for validation. 
Four designs (swerv\_wrapper, tinyRocket, dynamic\_node
and MegaBoom) are excluded from training and used
for unseen-design evaluation.
We train {\em MLBuf} using the AdamW optimizer,
with a learning rate of $3 \times 10^{-4}$, weight
decay of $10^{-2}$, and gradient-norm clipping at 1.0.
The learning rate is linearly increased during the
first 10\% of training steps (up to 500 steps) and
then decreased using cosine decay.
On the validation set, {\em MLBuf} achieves a gate
F1 score of 0.985, buffer-type accuracy of 0.82 for
matched buffers, mean absolute count error of 0.15
buffers per level, and median location error of
17\,$\mu m$.
We use GELU~\cite{HendrycksG2016} as the activation
function. The final model contains 0.64 million
parameters.

\begin{table}[ht]
\centering
\caption{Hyperparameters of MLBuf.}
\label{tab:MLBuf_hyperparameter}
\small
\begin{tabular}{l|l}
\hline
\textbf{Hyperparameter} & \textbf{Value} \\
\hline
Node Feature Dimension ($d_f$) & 15 \\
Net-context Feature Dimension ($d_c$) & 9 \\
Embedding Dimension ($d$) & 128 \\
Transformer Encoder Layers / Heads & 3 / 4 \\
Cluster Self-attention Layers / Heads & 1 / 4 \\
Dropout & 0.1 \\
Number of Clusters ($K$) & 32 \\
Gumbel-softmax Temperature ($\tau$) & 1.0 $\rightarrow$ 0.3 (12 epochs) \\
Gate Threshold & 0.40 \\
Maximum Recursion Depth & 5 \\
Buffer Type Output Dimension & 6 (one-hot) \\
Buffer Location Output Dimension & 2 (x, y) \\
\hline
\end{tabular}
\end{table}

\subsubsection{Ad-Hoc Baseline}
~\label{subsec_app:ad_hoc}
This is a rule-based analytical 
approximation used as a baseline for comparison,
as discussed in Section~\ref{subsec:ppa_val}. 
The approach involves (i) estimating the net wirelength 
using a wireload model (WLM)~\cite{CaldwellKMMZ99};
(ii) applying a pessimism margin of 25\% to 
the estimated wirelength to account for placement dynamics 
and congestion effects; (iii) estimating the buffer count by 
dividing the total load capacitance 
(wire + sinks) in one net by the max capacitance
limit of BUF\_X2;
(iv) adding an additional overhead factor of 20\% 
for routing detours and inefficiencies; and 
(v) allocating the estimated buffer area ``non-uniformly'' 
within the bounding box of the net.  
All hyperparameters (25\%, BUF\_X2, 20\%) are 
tuned to enhance the accuracy of estimated buffer 
counts by benchmarking against 
OR rsz, thereby improving
the competitiveness of this baseline.

\newpage
\section{Artifact Appendix}

\subsection{Abstract}
This appendix describes details for using {\em MLBuf} artifact.
It has passed MLCAD 2025 artifact evaluation and been verified
for reproducibility.
All data, codes and scripts are available at \cite{MLBuf}.

\subsection{Artifact check-list (meta-information)}

{\small
\begin{itemize}
  \item {\bf Algorithm: }Recursive learning, generative model, analytical global placement.
  \item {\bf Program: }Python (PyTorch), OpenROAD, OpenSTA.
  \item {\bf Compilation: }Build OpenROAD using the version provided in \cite{MLBuf}; Python compiler.
  \item {\bf Model: }{\em MLBuf}, a recursive learning-based generative buffering model.
  \item {\bf Data set: }Training dataset collected from OpenROAD rsz, publicly available at \cite{MLBuf}.
  \item {\bf Run-time environment: }Python 3 with all required libraries
  specified in the requirements; OpenROAD (directory OR\_branch\_integration/OpenROAD in \cite{MLBuf}, branch mlbuf-v2).
  \item {\bf Hardware: }AMD EPYC 7742 64-Core Processor CPU (128 threads); NVIDIA A100-SXM4-80GB GPU. 
  \item {\bf Execution: }Virtual buffering during global placement.
  \item {\bf Metrics: }Post-route wirelength, WNS, TNS, and power (with respect to TCP).
  \item {\bf Output: }Buffer-porosity-aware global placement results.
  \item {\bf Experiments: }Training the {\em MLBuf} model; integrating {\em MLBuf} into OpenROAD for {\em MLBuf-RePlAce}.
  \item {\bf Proprietary EDA tools: }Cadence Genus 21.1, Cadence Innovus 21.1.
  \item {\bf How much disk space required (approximately)?: }30GB (plus 20\,GB for the complete training data).
  \item {\bf How much time is needed to prepare workflow (approximately)?: }1 day.
  \item {\bf How much time is needed to complete experiments (approximately)?: }2-3 days.
  \item {\bf Publicly available?: }Yes.
  \item {\bf Code licenses (if publicly available)?: }BSD3 open-source license.
  \item {\bf Data licenses (if publicly available)?: }BSD3 open-source license.
  \item {\bf Workflow framework used?: }OpenROAD-flow-scripts.
\end{itemize}
}

\subsection{Description}

\subsubsection{How to access}
The repository, including datasets, source codes and scripts, is available at \cite{MLBuf}.
Detailed setup and usage instructions are provided in the README file.

\subsubsection{Hardware dependencies}
The experiments were conducted on a Linux server with an AMD EPYC 7742 64-Core Processor CPU (128 threads), 503GB RAM and an NVIDIA A100-SXM4-80GB GPU.

\subsubsection{Software dependencies}
Python 3 with all packages specified in the requirements file. 
OpenROAD built from the version in \cite{MLBuf}.

\subsubsection{Commercial software dependencies}
Cadence Genus 21.1 for synthesis; Cadence Innovus 21.1 for commercial‑flow validation.

\subsubsection{Data sets}
The training datasets were collected from OpenROAD rsz and are publicly available at \cite{MLBuf}.

\subsubsection{Models}
{\em MLBuf}, a recursive learning-based generative buffering model.

\subsection{Installation}
Detailed installation instructions are provided in the README file of the GitHub repository \cite{MLBuf}. These include Python environment setup, dependency installation, dataset, and execution scripts.

\subsection{Experiment workflow}
The complete workflow is described in the paper and 
the README file in the GitHub repository \cite{MLBuf}.
It includes training {\em MLBuf}, building OpenROAD from the specified version,
and conducting {\em MLBuf-RePlAce} to produce buffer-porosity-aware global placement results.

\subsection{Evaluation and expected results}
Evaluation results are provided in the paper, including
performance of predicted buffer-embedded trees and the post-route PPA results of {\em MLBuf-RePlAce}.

\subsection{Experiment customization}
Users can customize experiments by adjusting {\em MLBuf} hyperparameters such as
clustering output dimension and the number of clusters. 
Additionally, users can adjust the overflow\_list in the provided scripts, which 
controls the number of virtual buffering (i.e., ERC fixing in Figure~\ref{fig:overall_flow} of the paper) iterations performed during global placement.

\subsection{Methodology}

Submission, reviewing and badging methodology:

\begin{itemize}
  \item \url{https://www.acm.org/publications/policies/artifact-review-and-badging-current}
  \item \url{http://cTuning.org/ae/submission-20201122.html}
  \item \url{https://github.com/ml-eda/artifact-evaluation/}
\end{itemize}

\end{document}